\documentclass[letterpaper]{article} 
\usepackage{aaai25}  
\usepackage{times}  
\usepackage{helvet}  
\usepackage{courier}  
\usepackage[hyphens]{url}  
\usepackage{graphicx} 
\usepackage{natbib}  
\usepackage{caption} 
\usepackage{algorithm}

\usepackage{newfloat}
\usepackage{listings}
\DeclareCaptionStyle{ruled}{labelfont=normalfont,labelsep=colon,strut=off} 
\floatstyle{ruled}
\newfloat{listing}{tb}{lst}{}
\floatname{listing}{Listing}
\usepackage[utf8]{inputenc} 
\usepackage[T1]{fontenc}    
\usepackage{url}            
\usepackage{booktabs}       
\usepackage{amsfonts}       
\usepackage{nicefrac}       
\usepackage{microtype}      
\usepackage{xcolor}         
\usepackage{times}
\usepackage{soul}
\usepackage{url}
\usepackage{caption}
\usepackage{algpseudocode}
\usepackage{xpatch}
\makeatletter
\xpatchcmd{\algorithmic}{\ALG@tlm\z@}{\ALG@tlm\z@\leftmargin\z@}{}{}
\makeatother

\usepackage{amsmath}
\usepackage{amsthm}
\usepackage{booktabs}
\usepackage[switch]{lineno}
\usepackage{tabularx}
\usepackage{amsmath}
\usepackage{amsfonts}
\usepackage{subcaption}
\usepackage{amssymb}
\newtheorem{prop}{Proposition}

\title{TADA: Temporal Adversarial Data Augmentation for Time Series Data}
\author {
    Byeong Tak Lee,
    Joon-myoung Kwon,
    and Yong-Yeon Jo}
\affiliations {
    Medical AI Co. Ltd., Seoul, Republic of Korea \\
    \{bytaklee, cto, yy.jo\}@medicalai.com 
}

\usepackage{bibentry}

\begin{document}

\maketitle

\begin{abstract}

Domain generalization aim to train models to effectively perform on samples that are unseen and outside of the distribution. Adversarial data augmentation (ADA) is a widely used technique in domain generalization. It enhances the model robustness by including synthetic samples designed to simulate potential unseen scenarios into the training datasets, which is then used to train the model. However, in time series data, traditional ADA approaches often fail to address distribution shifts related to temporal characteristics. To address this limitation, we propose Temporal Adversarial Data Augmentation (TADA) for time series data, which incorporate time warping into ADA. Although time warping is inherently non-differentiable, ADA relies on generating samples through backpropagation. We resolve this issue by leveraging the duality between phase shifts in the frequency domain and time shifts in the time domain, thereby making the process differentiable. Our evaluations across various time series datasets demonstrate that TADA outperforms existing methods for domain generalization. In addition, using distribution visualization, we confirmed that the distribution shifts induced by TADA are clearly different from those induced by ADA, and together, they effectively simulate real-world distribution shifts.

\end{abstract}
%

\section{Introduction}

Most machine learning models are developed on the assumption that their training data reliably represents the broader population. 
Unfortunately, this assumption is often incorrect. The distribution of real-world data often deviates from that of the training dataset, and such discrepancies lead to performance degradation~\cite{volpi2018generalizing,zhao2020maximum,zhou2021domain,muandet2013domain}. 
An adversarial data augmentation (ADA) is one of the approaches that addresses distribution shift problem by solving worst-case scenarios around the original distribution~\cite{volpi2018generalizing, zhao2020maximum}. 
By training the model with adversarial synthetic samples, which are made extremely difficult for the model to classify, we ensure that the model performs reliably under less challenging conditions, such as distribution shift.

Time series data can be depicted on a two-dimensional space: the amplitude axis (y-axis) representing the magnitude of data points and the temporal axis (x-axis) indicating the sequence and timing of these points. 
While ADA effectively simulates samples that reflect distribution shifts related to the amplitude axis, it is limited in generating samples that address distribution shifts along the temporal axis~\cite{chatfield2013analysis,shumway2000time}. 
Given that distribution shifts in time series data are associated with both axes, it is crucial to develop methods that also account for changes along the temporal axis. 

In order to address the challenge of distribution shift along the temporal axis, we propose Temporal Adversarial Data Augmentation (TADA) for time series data, which incorporates a time warping technique into ADA process~\cite{muller2007dynamic,yang2021time}.
However, incorporating time warping into ADA presents a significant challenge. Specifically, the implementation of backpropagation is impeded by the non-differentiable nature of the index mapping used in time warping, which is a critical component in generation of adversarial sample in ADA. 
To address this problem, we propose a differentiable time warping approach by leveraging the duality between the non-differentiable time shifts, a fundamental unit of time warping, and the differentiable phase shifts.

We used various time series datasets, including electrocardiogram (ECG), electroencephalogram (EEG), and human activity recording (HAR)~\cite{reyna2021will,gagnon2022woods} to evaluate the effectiveness of our method for distribution shifts. Each dataset consists of multiple distinct domains. 
We conduct experiments in a single source domain generalization scenario, where models are trained on a single domain and then evaluated on the other remaining domains~\cite{wang2021learning}.

The results clearly demonstrate that TADA enhances model robustness across target domains with different distributions. Moreover, the combination of TADA and ADA, which perturb signals in both temporal and amplitude axes, consistently achieves the best performance in average across three type of datasets.
Implementing TADA is straightforward that requires inserting a differentiable time-warping block between the input and the encoder. This ensures compatibility with pre-existing ADA enhancement techniques that modifies loss function of ADA. When we applied pre-existing ADA enhancement techniques to TADA, we found a substantial improvement in performance.
Finally, by visualizing the representation with UMAP~\cite{mcinnes2018umap}, we confirm that TADA effectively simulates real-world distributions in latent space, emphasizing its value in addressing distribution shift problems in time series data.

The contributions of this paper are as follows: (1) We identify a significant limitation in existing ADA approaches for time series data, particularly their inability to effectively handle temporal dynamics and the the associated distribution shifts. (2) We propose a novel ADA method specifically designed to perturb data along the temporal axis, thereby enhancing the robustness of models against time-related variations in data distribution. (3) We demonstrate the effectiveness of our proposed method through extensive experiments within the single-source domain generalization framework.


    

\section{Related works}
\paragraph{Domain Generalization.} 
The distribution shift between source domain and unseen target domain often lead to degradation in performance~\cite{recht2019imagenet, moreno2012unifying, taori2020measuring}.
Domain generalization tackles these problems, aiming to train models that effectively generalize across multiple unseen target domain~\cite{blanchard2011generalizing,zhou2021domain,shu2021open,zhou2022domain}.

Depending on the coverage of the source domain, domain generalization is categorized into single-source and multiple-source domain generalization~\cite{wang2021learning,zhou2022domain}. 
In single-source domain generalization, the objective is to generalize the model when only a single domain is available for training. Since identifying domain-specific characteristics within the training dataset is impossible, single-domain generalization relies on generating fictitious distribution that simulate unseen domains, such as data augmentation-based methods~\cite{volpi2018generalizing,zhao2020maximum,zhou2021domain,wang2021learning}. 
On the other hand, the multiple-source domain generalization aims to train models to learn invariant features across multiple domains, ensuring they remain unaffected by domain shift in the source domain.
This is based on the premise that the invariant characteristics identified in the source domains will consistent in unseen target domain. Domain alignment, which seeks to these invariant features, is a widely adopted method in in multiple-source domain generalization~\cite{muandet2013domain,wang2021respecting}. 

\paragraph{Adversarial Data Augmentation.} 

Adversarial Data Augmentation (ADA) improves a model's ability to generalize across different domains or unseen distributions by solving the worst-case problem, as described in Equation~\ref{eq:ada}~\cite{volpi2018generalizing}. 
In particular, a model is trained to maintain robust performance in fictitious shifted distribution $P$ (i.e., domain shift scenarios), which may differ from the original distribution $P_0$. 
\begin{equation}
    \text{min}_{\theta \in \Theta} 
    \{ 
    \text{sup}_P
    \{ \mathbb{E}[\mathcal{L}(\theta;X,Y)]-\gamma D_\theta(P, P_0) \}  
    \},
\label{eq:ada}
\end{equation}
where $X$ and $Y$ denote a sample and its corresponding label, respectively, and $\gamma$ is a penalty parameter.

In practical, an iterative training procedure is employed that alternates between a maximization phase and a minimization phase.
During the maximization phase, data points are generated from fictitious distribution $P$, which is constrained to deviate within a distance $P$ from the source distribution $P_0$. 
On the other hand, in the minimization phase, the network $\theta$ is updated to minimize loss on the data sampled from both $P$ and $P_0$.

Several studies have been proposed to enhance the effectiveness of of ADA. Zhao et al.~\cite{zhao2020maximum} incorporated a term to maximizing the entropy of feature representation term in the loss function, which leads to generated samples covering a broad spectrum of potential distribution shifts. 
On the other hand, Qiao et al.~\cite{wang2021learning} introduced a the meta-learning scheme and the Wasserstein Auto-Encoder to relax the worst-case constraint.

\paragraph{Time Warping.}
Time warping is a technique utilized to align two sequence or time series data that may vary in speed. It is also employed as a data augmentation~\cite{muller2007dynamic,yang2021time}.
Time warping data augmentation generates a perturbed signal where the local speed of the signal is manipulated through index mapping using a warping path. 
Warping path represents a collection of time shift applied to each index in a sequence. Using this path, index mapping reorder the sequence into new positions in the warped sequence, resulting in either stretching or compressing.

The warping path must adhere to specific conditions~\cite{laperre2020dynamic}: (1) Maintaining Monotonicity: The ordering of points in the sequence is kept monotonic. (2) Boundary Alignment: The sequences are aligned at both the starting and ending points. (3) Warping Distance Constraint: The index shift caused by index mapping is limited, ensuring that the difference between corresponding indices does not exceed a predefined width. 

\section{Method}

\subsection{Problem Formulation}
ADA consists of maximization phase and a minimization phase, which focuses on data generations and updating the model, respectively. In order to generate samples with distribution shift in temporal axis, we redesign the maximization phase of ADA.
Equation~\ref{eq:x_update} represents the objective of the maximization phase from Equation~\ref{eq:ada}.
\begin{equation}
\begin{aligned}
    X^k \in \text{arg max}_{X} \bigl\{ 
    & \mathcal{L}(\theta ;(X^{k-1},Y)) \\
    & -\gamma c_\theta((X^{k-1},Y),(X^0,Y)) \bigr\}, 
\end{aligned}
\label{eq:x_update}
\end{equation}
where $k$ refers the number of iteration for adversarial update, $X^0$ and $X^k$ are original samples and adversarial perturbed samples at the iteration $k$, respectively. $\gamma$ is a penalty parameter, and $c_\theta$ is a distance  measure on the semantic space. Here, we use Wasserstein distance.

The maximization phase can simply be understood as adding adversarial noise to the input, expressed as $\hat{x} = x + \Phi$, where $\Phi$ represents the noise added. This process can be thought as passing the input sample through a function $F$: $\hat{x} = F(x;\Phi) = x + \Phi$.
Now, we can iteratively update $\Phi$ instead of $X$ as following Equation~\ref{eq:phi_update}:
\begin{equation}
\begin{aligned}
    \Phi^{k} \in \text{arg max}_{\Phi} 
    \bigl\{ 
    & \mathcal{L}(\theta ;(F(X;\Phi^{k-1}),Y)) \\
    & - \gamma c_\theta((F(X;\Phi^{k-1}),Y),(X,Y))    \bigl\}
\end{aligned}
\label{eq:phi_update}
\end{equation}

Our objective is to design a function $F$ that alters the temporal characteristic of the time series data $X$ without affecting its amplitude. 
To achieve this, we consider using a time warping function, which dynamically adjust the timing of sequence, as a function $F$. However, the time warping poses a challenge due to its non-differentiable nature of index mapping, which impedes the adversarial update of $\Phi$.

\subsection{Differentiable Time Warping}
To address this issue, we propose a new time warping function $F$ that operates equivalent to traditional time warping but is differentiable.
Fundamentally, we leverage the duality between a time shift in the time domain, which is a core unit of time warping and non-differentiable, and a phase shift in the frequency domain, which is differentiable.

We begin by presenting the mathematical formulation of time warping. 
Time warping is performed based on a specific warping path $\Delta = \{\delta_1, \delta_2, ..., \delta_N \}$, where each $\delta$ indicates a displacement from its original position. This mapping shifts the indices of the original sequence to a transformed sequence.
Applying time warping to the time series data $X=\{x_1, x_2, ..., x_N\}$ results in the transformed sequence $X'=\{x_{1+\delta_1}, x_{2+\delta_2}, ..., x_{N+\delta_N}\}$.

Now, we will see the time warping from an alternative perspective. 
We first consider splitting the sequence of original time series $X$ into overlapped segments $s_i \in S$, each represented by $\{x_{i-m}...,x_i,...,x_{i+m}\}$. 
Here, shifting the points of a segment $s_i$ by $\delta_i$ results in $\{x_{i-m+\delta_i}, \ldots, x_{i+\delta_i}, \ldots, x_{i+m+\delta_i}\}$. 
After shifting all segments by their respective $\delta_i$, we extract the central point of each segment $x_{i+\delta_i}$ to aggregate them. This results in the transformed sequence $X'=\{x_{1+\delta_1},x_{2+\delta_2},...,x_{n+\delta_N}\}$. 
This is identical to that obtained through time warping. 

A shifting points in this alternative perspective can be performed as a phase shift in the frequency domain. Unlike time shifts, which are non-differentiable, phase shifts are differentiable.
This allow us to replace the time shifts of each segment with phase shifts in the frequency domain..

We summarize this process in three steps: (1) We split the time series data \( X \) into multiple segments \( s \in S \) and transforms these segments into their corresponding outputs in the frequency domain. (2) We then apply the phase shift to the frequency domain components. (3) We transforms the phase-shifted components back into segments \( s \in S \) in the time domain. The time-shifted time series data \( X' \) is obtained by aggregating these segments \( S \). This process is formalized in the following Proposition~\ref{prop}.

\begin{prop}
\label{prop}

Applying time warping to a sequence is equivalent to applying phase shifts to the frequency components of overlapped sub-sequences of that sequence.
\begin{equation} 
\begin{aligned}
    & TimeWarping(X) \\
    & \ \ \ = \{ f'^{-1}(g(f(s_1),\Phi_1)) ,..., f'^{-1}(g(f(s_N),\Phi_N)) \},
\end{aligned}
\end{equation}
where $f(\cdot)$ and $f'^{-1}(\cdot)$ denotes Fourier and inverse Fourier transforms, respectively, and $g(\cdot,\Phi_i)$ represents the phase shifts.

\end{prop}
In the following, we will look into the details of each individual steps $f, g$, and $f'^{-1}$.

\subsubsection{Function $f$: Transformation to frequency domain}
\label{sec:3.2.1}
Function $f$ initially splits the time series data \( X \) into overlapped segments $S=\{s_1,...,s_N\}$. Each segment is defined by $s_i = \{x_{i-M}...,x_i,...,x_{i+M}\}$, where $M$ is a hyperparameter determining the size of each segment's window.
Subsequently, each segment $s_i$ is transformed into the frequency domain using the function \( f \) as the following Equation~\ref{eq:fft}:
\begin{equation}
    f(s_i)[k] = \sum_{n=0}^{N-1} s_i[n] \cdot e^{-j \frac{2\pi kn}{L}},
    \label{eq:fft}
\end{equation}
where $k$ is the frequency bin index and $N$ is the length sub-segment $s$.
In practice, this process can be implemented using the Short-Time Fourier Transform (STFT).
\begin{equation}
\label{eq:stft}
    \chi[m, k] = \sum_{n=-\infty}^{\infty} x[n]\cdot w[m-n] \cdot e^{-j\frac{2\pi kn}{L}}
\end{equation}
Here, the window function $w$ is defined as follows:
\begin{equation}
    w[m-n] = 
    \begin{cases}
    1, & \text{if } |m-n| < M,\\ 
    0, & \text{otherwise},
    \end{cases}
    \label{eq:window}
\end{equation}
where $M$ is the maximum window length.

\subsubsection{Function $g$: Time warping in frequency domain} 

Function $g$ perturbs the temporal characteristics of a segment by adding the noise that induces a phase shift, $g(f(s_i),\Phi_i) = f(s_i)+\Phi_i \cdot k$. 
Here, $\Phi \in \mathbb{R}^{N}$ corresponds to the warping path, where $N$ is the number of segments and also equals to the length of the original time series $X$.

The warping path $\Phi$ must satisfy several conditions~\cite{laperre2020dynamic}: (1) it must be monotonic, ensuring that the sequence advances in a single direction without reversals; (2) it must align with the boundaries, meaning that the start and end indices of the path should coincide with the start and end of the sequences being aligned; and (3) it must adhere to a warping distance constraint, which restricts the extent of the warping.

To meet these conditions, we impose several constraints on $\Phi$. This involves parameterizing $\Phi$ with the function  $h=h_3 \circ h_2 \circ h_1$ and a parameter $\phi$, where each $h_i$ function enforces one of the warping path conditions. Therefore, $\Phi$ is redefined as $\Phi = (h_3 \circ h_2 \circ h_1)(\phi)$, with $\phi$ representing the perturbation parameters. 
With this setup, the formula for temporal perturbation becomes $g(f(s_i),h(\phi_i)) = f(s_i)+ (h_3 \circ h_2 \circ h_1)(\phi_i) \cdot k$.

To begin with, we provide a explanation of function for maintaining monotonicity ($h_1$).
The monotonicity constraint ensures that the mapping of indices always progresses forward, never reversing direction. This means that if a point in the original sequence $a_i$ is matched with a point in a transformed sequence $b_j$, then $a_{i+1}$ can only be matched with $b_j$ or its subsequent points $b_{j+k}, k\geq 0$. This condition requires the difference between consecutive element of the warping path be non-negative.
To ensure that elements of the perturbation parameters $\phi$ remains non-negative, we subtract the minimum value found in the perturbation parameters $\phi$ from each element as follows:
\begin{equation}
\label{eq:cumsum}
    h_1(\phi_i) = \sum_{i=1}^{\tau} \phi_i - \text{min}(\phi)
\end{equation}

Next, we look into a description of function for boundary alignment ($h_2$).
To ensure that indices of elements in the transformed sequence align with the boundaries of the original sequence (i.e., starting and ending indices), 
we first normalize perturbation parameters $\phi_i$ and then scale the normalized path to match the length of original signal. 
The result of this process is actually a warping path. 
On this warping path, each element undergoes further transformation: it is subtracted by its own index order. This subtraction results in a value representing the difference between the original index and the corresponding index in the shifted sequence.
This is represented as below:
\begin{equation}
\label{eq:tada-normalization}
    h_2(\phi_i) = \frac{\text{max}(\phi) - \phi_i}{\text{max}(\phi) - \text{min}(\phi)} \cdot N - i ,
\end{equation}
where $i$ denotes a index, and $N$ denotes the total number of the segment (i.e. signal length).

Finally, we introduce a function for warping distance constraint ($h_3$).
To prevent the transformed sequence from deviating excessively from the original, a deviation constraint defined by $\phi_{\text{max}}$ is imposed as described in Equation~\ref{eqn:tada-clip}.
For instance, setting $\phi_\text{max}$ to 10 ensures that no index in the transformed sequence can deviate more than 10 indices away from its corresponding original index.
This approach keeps deviations within acceptable boundaries, thereby preserving the integrity of the transformation process.
\begin{equation}
\label{eqn:tada-clip}
    h_3(\phi_i) = \phi_i \cdot \min\left(\frac{\phi_{\text{max}}}{||\phi||_{\infty}}, 1\right) 
\end{equation}
In the equation above, if the infinity norm of the elements of $\phi$ (i.e., $||\phi||_{\infty}$) is greater $\phi_{\text{max}}$, $\phi$ is rescaled to ensure that all element absolute values remain below $\phi_{\text{max}}$.

\subsubsection{Function $f'^-1$: Reconstruction to time domain}
After perturbing the segments in the frequency domain, we obtain a perturbed frame represented as $g(f(s_i),h(\phi_i))$.
The function $f^{-1}$ then transforms each perturbed frame back to the time domain. Here, applying the inverse Fourier transformation to each frame is equivalent to perform the inverse STFT (ISTFT). Thus, we implemented ISTFT to convert a time frequency domain component $\chi'$ into the time domain, as expressed in Equation~\ref{eq:istft}.
\begin{equation}
    x'[n] = \sum_{m} \bigg( \sum_{k=0}^{L-1} \chi'[m,k] \cdot e^{j\frac{2\pi kn}{L}}\bigg)\cdot w'[n - mM]
    \label{eq:istft}
\end{equation}
Here, we use the following window function:
\begin{equation}
\label{eq:window-istft}
    w'[n-m] = 
    \begin{cases}
    1, n=m\\ 
    0, \text{otherwise}
    \end{cases}
\end{equation}
Note that the window function, $w'$, is different to Equation~\ref{eq:window}.
Instead of averaging the overlapping windows in the function $f$ as described in Equation~\ref{eq:stft}, we extract the central value from each transformed segment using a specific window function $w'$.
This approach ensures reconstruction of the original signal from its perturbed segments without the interference caused by overlapping during aggregation.

\subsection{Training Procedure}

We describe the full procedure in Algorithm~\ref{alg:ada}. 
The overall training scheme is similar to that of ADA~\cite{volpi2018generalizing}, with the key difference in how adversarial samples are generated during maximization phase. 

In the minimization phase (line 5-8), the model $\theta$ is prepared for the maximization phase, enabling it to generate more challenging samples. The model is updated by minimizing the loss function $\mathcal{L}$ over $T_{min}$ iterations.

In the maximization phase (line 9-15), the dataset $D$ is expanded by incorporating adversarial samples. 
The current model $\theta$ derives a challenging perturbation parameter $\phi_i$ for each sample $(X_i,Y_i)$ over iteration $T_{max}$. 
The samples $(X’,Y’)$ augmented by applying the perturbation parameters $\phi$ are then appended to the dataset.

These alternated process is repeated over $K$ iterations, resulting in the updated dataset $D_K$. After completing alternated process, the models is trained for $T$ epochs using the updated dataset.

\begin{algorithm}
\small
\caption{Training Procedure using Temporal Adversarial Data Augmentation}

\label{alg:ada}

\begin{algorithmic}[1]

\State \textbf{Input:} original dataset $D_0 = \{X_i, Y_i\}_{i=1}^{n}$ and initialized weights $\theta_0$, perturbation parameter $\phi_0$
\State \textbf{Output:} learned weights $\theta$

\State \textbf{Initialize}: $\theta \gets \theta_0$
\For{$k = 1, ..., K$}
    \For{$t = 1, ..., T_{min}$} \hspace*{50pt} $\triangleright$ {Minimization phase}
        \State Sample $(X, Y)$ uniformly from dataset $D_{k-1}$
        \State $\theta \gets \theta - \alpha \nabla_\theta \mathcal{L}(\theta; X, Y))$
    \EndFor    
        \For{\textbf{all} $(X_i,Y_i)\in D_{0}$} \hspace*{43pt}$\triangleright$ {Maximization phase}

            \State \textbf{Initialize}: $\phi_{i}^k \gets \phi_0$
            
            \For{$t = 1,...,T_{max}$}
                \State $\phi_{i}^k \gets \phi_{i}^k + \eta \nabla_\phi \{ \mathcal{L}(\theta;(F(X_{i},\phi_{i}^k),Y_i))$ 
                \Statex \hspace{110pt} $- \gamma c_\phi((F(X_{i},\phi_{i}^k),Y_i),(X_i,Y_i))\} $
            \EndFor
            \State Expand dataset $D_{k}$ by appending $(F(X_{i},\phi_{i}^k) , Y_{i})$
            \EndFor
\EndFor

\For{$t = 1,...,T$}
    \State Sample $(X, Y )$ uniformly from dataset $D_{K}$
    \State $\theta \gets \theta - \alpha \nabla_\theta \mathcal{L}(\theta; (X, Y))$
\EndFor
\end{algorithmic}
\label{alg:ada}
\end{algorithm}

\section{Experiments}

We evaluate our proposed method against state-of-the-art approaches in the single domain generalization framework, which involves training models on a single domain and assessing their performance on other domains. Additionally, we conduct ablation studies to assess the compatibility of the proposed model with other widely used approaches. Lastly, by analyzing the representations extracted from models trained using our methodology, we verify whether our method effectively simulates a valid distribution shift.

\subsection{Experimental Setup}
\paragraph{Datasets.} 
We used three type of datasets with several sub-datasets as shown in Table~\ref{tab:dataset}. 
These sub-datasets were sourced from distinct places or collected using different devices, thereby presenting inherent distribution differences. 

\textsf{Physionet Challenge 2021 (Physionet) dataset} \cite{reyna2021will} is a public dataset for electrocardiogram (ECG) research. Among seven sub-datasets of Physionet datasets, we selected five sub-datasets that each contain over 1,000 ECGs: PTBXL \cite{wagner2020ptb}, Chapman-Shaoxing (Chapm) \cite{zheng202012}, Ningbo \cite{zheng2020optimal}, G12EC, and CPSC2018 \cite{liu2018open}. 
Each dataset has unique demographic profiles: PTBXL was gathered in Germany, Chapm, Ninbgo, CPSC2018 were collected in China, and G12EC was collected in the USA.
While the datasets involve the 26 different types of arrhythmia class, only seven of these types are present in all sub-datasets. Therefore, we focus on the multi-label classification of these seven specific classes.

\textsf{Woods-PCL (PCL) dataset}~\cite{gagnon2022woods} is a electroencephalogram (EEG) collection.
It was collected from three distinct medical centers, each with different collection procedures from different research group. 
These difference show how the human intervention in data gathering process can contribute to the distribution shift~\cite{koh2021wilds}.
The task of the dataset is to classify imagery hand movements.

\textsf{Woods-HHAR (HHAR) dataset}~\cite{gagnon2022woods} is a signal dataset that records human activity recognition from smart devices qeuipped with three-axis accelerometers and gyroscopes. 
Data was collected from five different types of devices, each processed using distinct hardware features and signal processing techniques. These variations can lead to distribution shift between the different smart devices.
The objective of the task is classify one of six human activity, including sitting, walking, and running. 

\begin{table}[h]
\centering
\scriptsize
\caption{Datasets used in the experiments.}
\begin{tabular}{l l l l }
\toprule
\textbf{} & \textbf{Physionet} & \textbf{PCL} & \textbf{HHAR} \\
\midrule
\textbf{Signal} & ECG & EEG & Accelero. and gyro. \\
\textbf{Dist. shift} & Ethnicity & Human intervention & Device \\
\textbf{Task} & Arrhythmia & Hand movement & Human activity \\
\textbf{Num. of class} & 7 & 2 & 6 \\
\textbf{Num. of sample} & 87,662 & 22,597 & 13,673 \\
\bottomrule
\end{tabular}
\label{tab:dataset}
\end{table}

\paragraph{Implementation details.}
Previous studies have demonstrated that shallow neural networks can achieve performance comparable to or even better than deeper networks in signal analysis~\cite{kanani2020shallow, lee2023optimizing}. 
Based on these findings, we focus on experiments using relatively shallow ResNet18 as our backbone network, following the previous research.
For our experiments, we used the Adam optimizer~\cite{kingma2014adam} and the cosine annealing scheduler~\cite{smith2019super} with a peak epoch of 10. The experiments were run with a batch size of 512. 
The dropout rate is randomly chosen from $0$ to $0.3$ in increments of $0.05$. 
For the Physionet dataset, the learning rate and weight decay were randomly selected within the range of $[10^{-5}, 10^{-3}]$. 
In contrast, for the PCL and HHAR datasets, the learning rate and weight decay were randomly selected from the range of $[10^{-4}, 10^{-2}]$.

For hyperparameters related to TADA, we set both $T_{max}$ and $T_{min}$ to 10, and $K$ to 2.
The window size $M$ for STFT in TADA was fixed to 10. 
For both ADA and TADA, $\gamma$ was randomly selected in $\{0.1, 1, 10\}$. 
The parameter $\eta$ was selected from $\{100, 200, 500\}$ for ADA, while it was fixed as $1$ for TADA. 

As a scheduler for the hyperparemeter tuning, we used the asynchronous successive halving algorithm, set with a grace period 10 and a reduction factor of 2~\cite{li2018massively}. 
The entire training workflow was implemented with the Ray framework~\cite{moritz2018ray}. 

\paragraph{Comparison methods.}
We compared the proposed method with six approaches:
(1) \textsf{Empirical Risk Minimization (ERM)} is a fundamental training principle widely used in machine learning. This is used as a standard baseline method.
(2) \textsf{Mixup}~\cite{zhang2017mixup} uses convex combinations of pairs of samples and their labels as inputs to train a network. 
(3) \textsf{RandConv}~\cite{xu2020robust} uses multi-scale random convolutions as a data augmentation method to generate samples with randomized local textures while preserving global shapes.
(4) \textsf{ADA}~\cite{volpi2018generalizing} is an adversarial training to generate new training examples by introducing perturbations designed to maximize prediction error of the model. 
(5) \textsf{ME-ADA}~\cite{zhao2020maximum} is an extension of ADA by adopting a maximum-entropy criterion.
(6) \textsf{M-ADA}~\cite{qiao2020learning} is an extension of ADA that relax the worst-case constraint to encourage large domain shift.

\paragraph{Evaluation.}
To ensure a balanced assessment across classes, we used the macro F1 score as the performance metric. 
We evaluated the proposed method on the single domain generalization framework. Thus, we trained the model on each sub-dataset and then evaluated it on the remaining sub-datasets, and the average F1 score across these sub-datasets are used for comparison.
It is known that experiments lacking rigorous hyperparameter tuning often result in misleading conclusions about the efficacy of a method, potentially overstating its actual performance~\cite{arnold2023role, yang2020hyperparameter}.
In order to prevent this, we meticulously tuned hyperparameters for each training by exploring 30 different combinations.

\subsection{Performance on Datasets with Different Domains}
In the tables below, scores in boldface indicate the best performance, and underlined scores represent the second best. Additionally, we have included the results for a combined version of ADA and TADA, referred to as \textsf{TADA+}, in our experiments.

\paragraph{Physionet dataset.}
Table~\ref{tab:result-ecg} shows the average F1 scores on the Physionet dataset.
For the PTBXL dataset, our proposed methods, including TADA and TADA+, were the most effective. In G12EC dataset, the performance among tested methods were very similar to each other, with TADA leading slightly. However, the CPSC2018 dataset, which generally presents lower scores, identified Mixup to be the most effective method. For the Chapm dataset, TADA+ achieved the best result. Similarly, in the Ningbo dataset, TADA+ outperformed other methods. 

While the average scores of TADA+ generally rank among the top, the average performance of ADA and its variations, including ADA, ME-ADA and M-ADA, was even lower than that of ERM. 
These results highlights the limitation of perturbation solely focusing on amplitude and the effectiveness of incorporating perturbation on temporal characteristics in ECG datasets.

\begin{table}[h]
\centering
\scriptsize
\caption{F1 scores on the Physionet dataset}
\begin{tabular}{l c c c c c c}
\toprule
\textbf{Method} & \textbf{PTBXL} & \textbf{G12EC} & \textbf{CPSC} & \textbf{Chapm} & \textbf{Ningbo} & \textbf{Average} \\
\midrule
\textbf{ERM} & 0.4943 & 0.4628 & 0.3820 & 0.4561 & 0.5174 & 0.4625 \\
\textbf{Mixup} & 0.4918 & \underline{0.4723} & \textbf{0.3971} & 0.4554 & {0.5321} & \underline{0.4697} \\
\textbf{RandConv} & 0.4863 & 0.4609 & \underline{0.3938} & \underline{0.4702} & \underline{0.5349} & 0.4692 \\
\textbf{ADA} & 0.4681 & 0.4631 & 0.3749 & 0.4561 & 0.5149 & 0.4554 \\
\textbf{ME-ADA} & 0.4835 & 0.4578 & 0.3752 & {0.4692} & 0.5276 & 0.4627 \\
\textbf{M-ADA} & 0.4769 & 0.4694 & 0.3719 & 0.4649 & 0.5278 & 0.4622 \\
\textbf{TADA} & \textbf{0.4957} & \textbf{0.4745} & 0.3565 & 0.4588 & 0.5272 & 0.4625 \\
\textbf{TADA+} & \textbf{0.4957} & 0.4721 & 0.3737 & \textbf{0.4737} & \textbf{0.5389} & \textbf{0.4708} \\
\bottomrule
\end{tabular}
\label{tab:result-ecg}
\end{table}

\paragraph{PCL dataset.}
Table~\ref{tab:result-eeg} presents the average F1 scores on the PCL dataset. 
In Dataset1, RandConv emerged as the best method, with TADA following as the second best. In both Dataset2 and Dataset3, TADA+ achieved the best performance. 
Overall, TADA+ demonstrated the superior performance in the EEG domain.

In contrast to the result in Physionet dataset, we found that ADA slightly surpass ERM in average performance. However, the ADA variants, ME-ADA and M-ADA, performed worse than both ERM and ADA, showing that the variants of ADA do not always lead the performance enhancement. While TADA alone did not yield improvements over ERM, the combination of ADA with TADA led to a significant improvement in results. These findings suggest that amplitude perturbation are effective in EEG dataset, but combining with temporal perturbation can lead further performance improvement.

\begin{table}[h]
\centering
\scriptsize
\caption{F1 scores on the PCL dataset}
\begin{tabular}{l c c c c}
\toprule
\textbf{Method}       & \textbf{Data1} & \textbf{Data2} & \textbf{Data3} & \textbf{Average} \\
\midrule
\textbf{ERM}       & 0.6367 & 0.6080 & 0.5898 & 0.6115 \\
\textbf{Mixup}     & 0.6304 & 0.5940 & \underline{0.6001} & 0.6082 \\
\textbf{RandConv}  & \textbf{0.6877} & 0.5875 & 0.5732 & \underline{0.6161} \\
\textbf{ADA}       & 0.6368 & \underline{0.6158} & 0.5935 & 0.6154 \\
\textbf{ME-ADA}    & 0.6265 & 0.5950 & 0.5895 & 0.6036 \\
\textbf{M-ADA}     & 0.6352 & 0.6053 & 0.5931 & 0.6112 \\
\textbf{TADA}      & \underline{0.6426} & 0.5994 & 0.5923 & 0.6114 \\
\textbf{TADA+}  & 0.6203 & \textbf{0.6161} & \textbf{0.6277} & \textbf{0.6214} \\
\bottomrule
\end{tabular}
\label{tab:result-eeg}
\end{table}

\paragraph{HHAR dataset.}
Table~\ref{tab:result-hhar} shows the experimental result on the HHAR dataset. 
In Dataset1, TADA+ achieved the best performance. However, we found best performance in ME-ADA in Dataset2, TADA in Dataset3, ERM in Dataset4, and Mixup in Dataset. The best-performing methods varied across different sub-datasets. 
Nevertheless, TADA+ ultimately achieves the best average performance.

Similar to the observations from EEG dataset, we found that ADA outperforms ERM, proving its effectiveness for domain generalization in human activity recognition dataset. However, the performance of both ME-ADA and M-ADA, which are designed to enhance ADA, were lower than that of ADA alone.
Interestingly, we found that TADA sometimes underperformed compared to ERM, which runs contrary to the trends seen in other datasets in the ECG and EEG domains. Although TADA had a small negative impact on performance, but when combined with ADA, there were significant enhancements. This synergistic effect of ADA and TADA was consistently observed throughout our experiments.

\begin{table}[h]
\centering
\scriptsize
\caption{F1 scores on the HHAR dataset}
\begin{tabular}{l c c c c c c}
\toprule
\textbf{Method}        & \textbf{Data1} & \textbf{Data2} & \textbf{Data3} & \textbf{Data4} & \textbf{Data5} & \textbf{Average} \\
\midrule
\textbf{ERM}       & 0.3441 & 0.4606 & \underline{0.6170} & \textbf{0.6164} & 0.6706 & 0.5417 \\
\textbf{Mixup}     & 0.3575 & 0.4213 & 0.6009 & 0.6023 & \textbf{0.6777} & 0.5319 \\
\textbf{RandConv}  & 0.4078 & 0.5202 & 0.5728 & 0.5888 & 0.6118 & 0.5405 \\
\textbf{ADA}       & 0.4594 & \underline{0.5206} & 0.5668 & 0.5663 & \underline{0.6711} & \underline{0.5568} \\
\textbf{ME-ADA}    & 0.3485 & \textbf{0.5357} & 0.5960 & 0.5873 & 0.6530 & 0.5441 \\
\textbf{M-ADA}     & 0.3739 & 0.4419 & 0.5774 & 0.5680 & 0.6639 & 0.5250 \\
\textbf{TADA}      & 0.3436 & 0.4479 & \textbf{0.6252} & 0.5898 & 0.6550 & 0.5323 \\
\textbf{TADA+}  & \textbf{0.5027} & 0.4869 & 0.5802 & 0.5822 & 0.6576 & \textbf{0.5619} \\
\bottomrule
\end{tabular}
\label{tab:result-hhar}
\end{table}

\subsection{Complementarity of TADA}
In practice, machine learning practitioners deploy various techniques together to improve performance. 
In this respect, the complementarity of the method is crucial, as it ensures that each technique contributes independently to the outcome without interfering or diminishing the effectiveness of others~\cite{sarker2021machine}.
Given its straightforward implementation, TADA can be readily combined with existing methods designed to enhance ADA. To evaluate the complementarity effect of TADA and TADA+, we applied techniques used to enhance ADA, such as M-ADA and ME-ADA, into both TADA and TADA+. This integration has resulted in variants such as ME-TADA, M-TADA, ME-TADA+, and M-TADA+

Table~\ref{tab:result-ablation} presents the average F1 scores for all domain datasets. 
Although applying ME and M to TADA and TADA+ does not consistently result in performance improvements, we observed a substantial enhancement in a few datasets.
Specifically, in the PCL dataset, M-TADA+ achieved a significant performance improvement over TADA+.
Furthermore, in the HHAR dataset, M-TADA led to considerable performance gain, and ME-TADA also improved performance compared to TADA alone.
These result align with the previous experiments, where we observed that the effectiveness of ME-ADA and M-ADA varies depending on dataset type. 
In summary, our finding confirm that TADA and TADA+ can be integrated well with existing methods for enhancing ADA and sometimes lead to substantial performance improvements.

\begin{table}[h]
\centering
\scriptsize
\caption{Average F1 score of the variants of ADA with TADA for all datasets}
\label{tab:result-ablation}
\begin{tabular}{l c c c c}
\toprule
\textbf{Method} & \textbf{Physionet} & \textbf{PCL} & \textbf{HHAR} \\
\midrule
\textbf{TADA}               & 0.4625 & 0.6114 & 0.5323 \\
\textbf{ME-TADA}           & 0.4542 & 0.6203 & {0.5492} \\
\textbf{M-TADA}             & {0.4650} & 0.6191 & {0.5878} \\
\midrule
\textbf{TADA+}           &  {0.4708} & {0.6214} & {0.5619} \\
\textbf{ME-TADA+}       & {0.4645} & {0.6211} & 0.5429 \\
\textbf{M-TADA+}       & 0.4598 & {0.6735} & 0.5268 \\
\bottomrule
\end{tabular}
\label{tab:result-ablation}
\end{table}

\begin{figure*}[h]
    \centering
    \hspace{0.5cm}
    \begin{subfigure}{0.32\textwidth}
        \includegraphics[width=4.6cm]{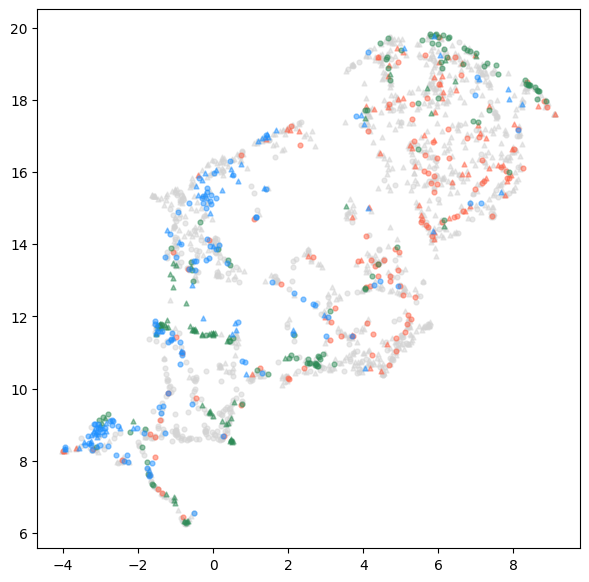}
        \caption{Physionet dataset}
        \label{fig:umap-ecg}
    \end{subfigure}
    \hspace{-0.3cm}
    \begin{subfigure}{0.32\textwidth}
        \includegraphics[width=4.6cm]{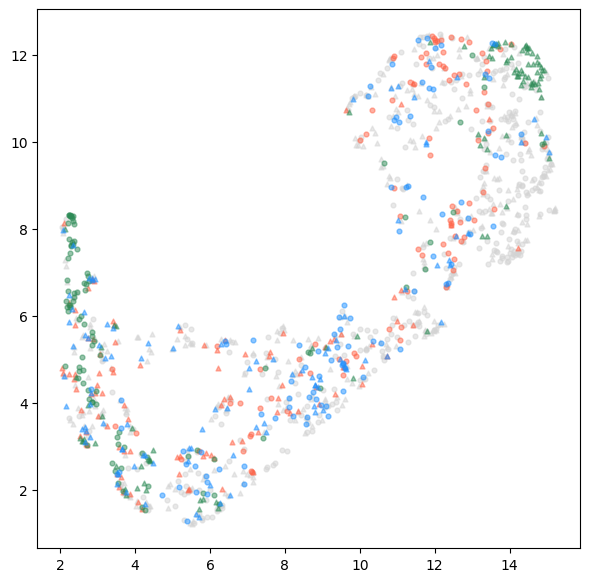}
        \caption{PCL dataset}
        \label{fig:umap-eeg}
    \end{subfigure}
    \hspace{-0.45cm}
    \begin{subfigure}{0.32\textwidth}
        \includegraphics[width=4.7cm]{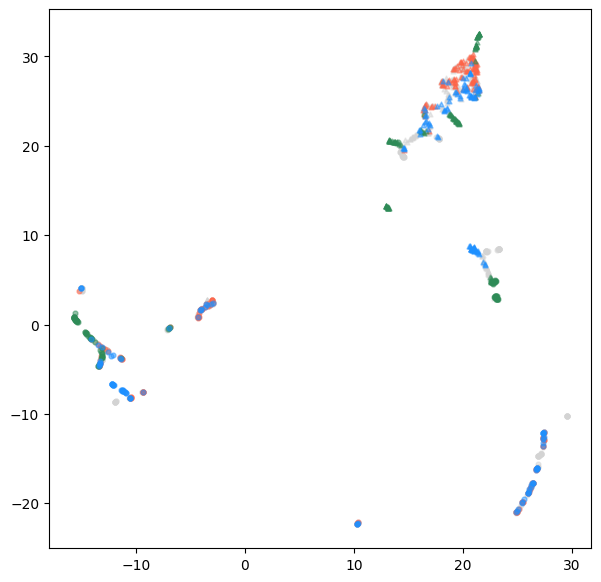}
        \caption{HHAR datset}
        \label{fig:umap-hhar}
    \end{subfigure}
    \caption{UMAP visualization for the representations of the datasets. \textcolor{orange}{Orange dots} represent the training samples, \textcolor{gray}{grey points} depict the unseen samples, \textcolor{blue}{blue dots} correspond to the representations augmented by ADA, and \textcolor{green}{green dots} represent those generated by TADA.}
    \label{fig:dist}
\end{figure*}

\subsection{Visualization of Effect of TADA}

\label{sec:4.4}
Our study hypothesizes that TADA introduces a unique type of distribution shift, caused by temporal perturbation, which is different from the amplitude perturbation induced by ADA. 
We expected that these two types of adversarial data augmentation, ADA and TADA, effectively simulate different types of distribution shifts in time series data. 
To verify this, we use a model that has been trained on one sub-dataset and then compare the representations from this sub-datasets with those generated through ADA and TADA. 
We visualized these representations using UMAP~\cite{mcinnes2018umap}.

Figure~\ref{fig:dist} displays a UMAP visualization of the datasets. 
Orange dots represent samples from the source sub-dataset, indicating the training data points.
For source dataset, we used G12EC for the Physionet dataset and Dataset1 for both the PCL and HHAR datasets.
Gray dots represents samples from external sub-datasets, indicating unseen samples. 
The noticeable difference in dot coverage between the gray and orange dots is observed, which demonstrates the distribution disparities between source and external datasets.
Blue dots show samples augmented by ADA, incorporating amplitude variations of the source data, while green dots represent ones generated by TADA, which specifically introduce temporal shifts.
The clear distinction between blue and green dots on the distribution map highlight the different types of distribution shifts each method generates. 
Moreover, these augmented points serve to fill the gap between the coverage of the source and target datasets, demonstrating an effective bridging of the distribution disparities.


\section{Conclusion}
We introduced the TADA to address domain generalization challenges specific to time series data. 
Conventional ADA approaches primarily focuses on addressing amplitude-related augmentations for dealing with distribution shifts and overlooks to address the distribution shifts associated with temporal characteristics, which are crucial for time series data.
By leveraging a differentiable time warping technique that utilizes the duality between time shift in the time domain and phase shifts in the frequency domain, TADA augments samples with temporal changes.
For extensive evaluations on the single domain generalization framework, we used various datasets with different domains including electrocardiograms, electroencephalograms, and human activity recordings.
On this experimental setup, we demonstrated that TADA or combination of TADA and ADA outperforms existing ADA alone and its variants.
In addition, TADA can be integrated seamlessly into existing ADA variations as a complementary method, and we observed a performance improvement when they are used together. This verified that they enhance their practice effectiveness, especially when multiple techniques are often deployed together.
Moreover, our analysis confirms that both ADA not only address different changes in data distribution but also successfully cover the distribution gap between the source and external datasets.



\bibliography{aaai25}

\end{document}